\documentclass[10pt,twocolumn,letterpaper]{article}

\usepackage[pagenumbers]{cvpr} % arXiv version

\usepackage{multirow}
\usepackage{subcaption}
\usepackage{microtype}
\usepackage{algorithm}
\usepackage{algorithmic}
\usepackage{pifont}

\newcommand{\bench}{\textsc{AgentComm-Bench}\xspace}
\newcommand{\ours}{\textsc{ResilientComm}\xspace}

\definecolor{cvprblue}{rgb}{0.21,0.49,0.74}
\usepackage[pagebackref,breaklinks,colorlinks,allcolors=cvprblue]{hyperref}

\def\paperID{*****}
\def\confName{CVPR}
\def\confYear{2026}

\title{AgentComm-Bench: Stress-Testing Cooperative Embodied AI\\Under Latency, Packet Loss, and Bandwidth Collapse}

\author{Aayam Bansal, Ishaan Gangwani\\
Synthetic Sciences\\
\texttt{\{aayam, ishaan\}@syntheticsciences.ai}}

\begin{document}
\maketitle
\begin{abstract}
Cooperative multi-agent methods for embodied AI are almost universally evaluated under idealized communication: zero latency, no packet loss, and unlimited bandwidth. Real-world deployment---on robots with wireless links, autonomous vehicles on congested networks, or drone swarms in contested spectrum---offers no such guarantees. We introduce \bench, a benchmark suite and evaluation protocol that systematically stress-tests cooperative embodied AI under six communication impairment dimensions: latency, packet loss, bandwidth collapse, asynchronous updates, stale memory, and conflicting sensor evidence. \bench spans three task families---cooperative perception, multi-agent waypoint navigation, and cooperative zone search---and evaluates five communication strategies, including a lightweight method we propose called \ours that applies redundant message coding with staleness-aware fusion. Our experiments reveal that communication-dependent tasks degrade catastrophically: stale memory and bandwidth collapse cause $>$96\% performance drops in navigation, while content corruption (stale/conflicting data) destroys perception F1 by $>$85\%. Critically, vulnerability depends on the interaction between impairment type and task design---perception fusion is immune to packet loss but amplifies corrupted data. Redundant message coding more than doubles navigation performance under 80\% packet loss. We release \bench as a practical evaluation protocol and recommend that all cooperative embodied AI papers report performance under at least three impairment conditions.
\end{abstract}

\section{Introduction}
\label{sec:intro}

Cooperative multi-agent systems---from autonomous vehicles sharing perception~\cite{xu2022opv2v,wang2020v2vnet} to robot teams coordinating exploration~\cite{yu2023asynchronous}---depend critically on inter-agent communication. The past five years have produced a rich landscape of learned communication protocols~\cite{sukhbaatar2016commnet,das2019tarmac,singh2019ic3net}, communication-efficient methods~\cite{liu2020when2com,hu2022where2comm,kim2019learning}, and cooperative perception systems~\cite{xu2022v2xvit,li2021disconet,hu2023coca3d}. Yet nearly all of these methods are evaluated under an implicit assumption: \emph{communication is instantaneous, lossless, and unlimited}.

This assumption is unrealistic. Wireless channels between robots experience packet loss rates of 5--30\%~\cite{farjam2023distributed}, latencies of 50--500\,ms, and bandwidth that fluctuates with interference and congestion. Vehicle-to-everything (V2X) links are subject to asynchronous clock domains~\cite{yu2022dairv2x}, stale information from processing delays~\cite{chamoun2025mappo}, and conflicting sensor readings across heterogeneous platforms. While some recent methods address individual failure modes---mmCooper~\cite{su2024mmcooper} handles delays in cooperative perception, V2XPnP~\cite{xu2023v2xpnp} addresses pose noise, and BVME~\cite{zhang2024bvme} applies variational bottlenecks for bandwidth constraints---no unified evaluation protocol examines the full spectrum of communication impairments.

We introduce \bench, a benchmark suite and evaluation protocol for stress-testing cooperative embodied AI under realistic communication failures. Our contributions are:

\begin{enumerate}[label=\textbullet]
    \item \textbf{A six-dimensional communication stress protocol} covering latency, packet loss, bandwidth collapse, asynchronous updates, stale memory, and conflicting sensor evidence, with parameterized severity levels enabling fine-grained robustness curves.
    \item \textbf{Three complementary task families}---cooperative perception, multi-agent waypoint navigation, and cooperative zone search---instantiated as lightweight grid-world simulations that isolate communication effects from perceptual complexity.
    \item \textbf{\ours}, a lightweight robust communication wrapper combining redundant message coding with staleness-aware fusion, demonstrating that simple engineering principles can improve robustness under packet loss.
    \item \textbf{A standardized evaluation protocol} including normalized performance drop, robustness curves, rank stability analysis, and failure mode taxonomy, which we recommend as a reporting standard for cooperative embodied AI.
\end{enumerate}

We position \bench as a practical evaluation protocol: the benchmark requires no specialized hardware, runs in under five minutes on a single CPU, and produces a comprehensive robustness profile for any cooperative method. The lightweight grid-world instantiation serves as a reference implementation; we encourage the community to apply the protocol to photorealistic simulators such as CARLA~\cite{dosovitskiy2017carla} or Habitat~3.0~\cite{puig2023habitat3} and to standard cooperative perception datasets~\cite{xu2022opv2v,yu2022dairv2x}.

\section{Related Work}
\label{sec:related}

\paragraph{Cooperative multi-agent communication.}
Learned communication protocols for multi-agent systems have evolved from broadcast-based architectures like CommNet~\cite{sukhbaatar2016commnet} to targeted messaging via attention~\cite{das2019tarmac} and gated communication~\cite{singh2019ic3net}. For cooperative perception, Who2com~\cite{liu2020who2com} and When2com~\cite{liu2020when2com} introduced handshake-based protocols that learn \emph{whom} and \emph{when} to communicate. V2VNet~\cite{wang2020v2vnet} established intermediate feature sharing as the dominant paradigm, further refined by DiscoNet~\cite{li2021disconet} with knowledge distillation and Where2comm~\cite{hu2022where2comm} with spatial confidence maps achieving 100,000$\times$ bandwidth reduction. Transformer-based fusion~\cite{xu2022v2xvit,xu2022cobevt} enables robust multi-agent feature aggregation. CoAlign~\cite{lu2023coalign} addresses pose errors, and CoCa3D~\cite{hu2023coca3d} demonstrates that collaborative cameras can rival LiDAR. Despite this progress, robustness to communication \emph{channel} failures remains largely unexamined.

\paragraph{Communication under realistic constraints.}
A recent survey~\cite{liu2025robust} identifies that most MARL methods assume instantaneous, reliable communication. Exceptions include SchedNet~\cite{kim2019learning}, which learns communication scheduling under bandwidth limits, ETCNet~\cite{hu2021event}, which applies event-triggered communication to reduce bandwidth occupancy, and IMAC~\cite{wang2020learning}, which uses information bottleneck principles for message compression. Li and Zhang~\cite{li2024context} propose context-aware personalized messages under bandwidth constraints. ET-MAPG~\cite{li2021etmapg} learns adaptive event triggers for communication, BVME~\cite{zhang2024bvme} applies variational bottlenecks for bandwidth-adaptive messaging, and CC-MADDPG~\cite{zhang2024ccmaddpg} uses loss-aware mutual information shaping. For cooperative perception specifically, mmCooper~\cite{su2024mmcooper} handles delays via confidence-guided multi-stage fusion, and V2XPnP~\cite{xu2023v2xpnp} addresses pose noise with temporal BEV compression. For asynchronous settings, ACE~\cite{yu2023asynchronous} extends MAPPO to handle action delays in multi-robot exploration, while Min~\etal\cite{min2023cooperative} provide convergence guarantees for asynchronous cooperative MARL. Zhou~\etal\cite{zhou2021multiagent} apply Bayesian belief updates to handle packet loss. However, these works each address a \emph{single} failure mode; no unified benchmark evaluates methods across the full spectrum of communication impairments.

\paragraph{Benchmarks for cooperative embodied AI.}
The cooperative perception community has produced several datasets: OPV2V~\cite{xu2022opv2v} (simulated V2V), V2X-Sim~\cite{li2022v2xsim} (multi-agent simulation), DAIR-V2X~\cite{yu2022dairv2x} (real-world vehicle-infrastructure), and V2V4Real~\cite{xu2023v2v4real} (real-world V2V). Photorealistic simulators including CARLA~\cite{dosovitskiy2017carla} provide controllable environments for V2X research. For embodied navigation, Habitat~3.0~\cite{puig2023habitat3} supports human-robot collaboration, and Stern~\etal\cite{stern2019mapf} formalize multi-agent pathfinding benchmarks. LLM-based cooperative agents~\cite{zhang2024coela,mandi2023roco} introduce new evaluation paradigms. V2XP-ASG~\cite{xiang2023v2xpasg} provides adversarial scene generation for V2X perception. Lin~\etal\cite{lin2020robustness} demonstrate that attacking a single agent's observations can collapse team performance. Yet none of these benchmarks systematically evaluate the effect of communication channel impairments on cooperative performance.

\section{The \bench Benchmark}
\label{sec:benchmark}

\begin{figure}[t]
\centering
\includegraphics[width=\columnwidth]{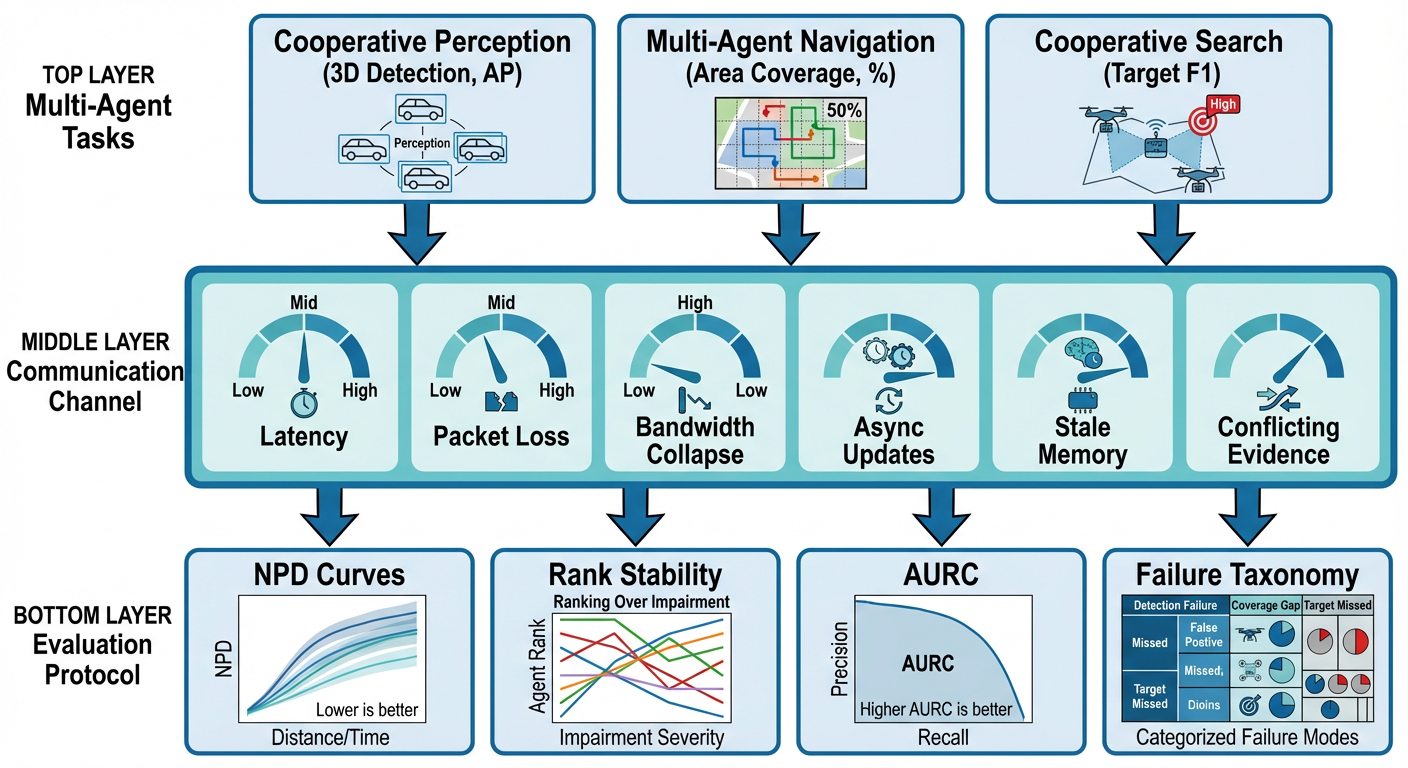}
\caption{\textbf{Overview of \bench.} The benchmark evaluates cooperative methods across three task families under six communication impairment dimensions, producing robustness profiles via standardized metrics.}
\label{fig:overview}
\end{figure}

\bench is designed around three principles: (1)~\emph{comprehensive coverage} of realistic communication failure modes, (2)~\emph{reproducible evaluation} requiring only lightweight simulation, and (3)~\emph{standardized reporting} via a fixed set of robustness metrics. Figure~\ref{fig:overview} provides an overview of the benchmark architecture.

\subsection{Communication Impairment Dimensions}
\label{sec:impairments}

We define six impairment dimensions, each parameterized by a scalar severity $\sigma \in [0, \sigma_{\max}]$. While latency, asynchrony, and stale memory share a temporal character, they differ in mechanism (transport delay vs.\ clock drift vs.\ refresh failure) and produce distinct degradation patterns (Appendix~\ref{app:orthogonal}):

\begin{enumerate}[label=\textbf{D\arabic*.},leftmargin=2em]
    \item \textbf{Latency} ($\sigma \in [0, 500]$ ms). Messages arrive with a fixed delay. At $\sigma{=}500$\,ms, agents operate on observations that are 10--15 decision steps old, a realistic scenario for cellular V2X links.
    \item \textbf{Packet loss} ($\sigma \in [0, 80]\%$). Each message is independently dropped with probability $\sigma/100$, following a Bernoulli model. At 80\% loss, agents receive on average only 1 in 5 messages.
    \item \textbf{Bandwidth collapse} ($\sigma \in [0, 100]\%$ reduction). The available channel capacity is reduced by $\sigma$\%, forcing methods that transmit large feature maps to either compress or drop information.
    \item \textbf{Asynchronous updates} ($\sigma \in [0, 10]$ steps). Agents operate on different clock domains, with each agent's state offset by a uniform random delay in $[0, \sigma]$ steps from the true environment state.
    \item \textbf{Stale memory} ($\sigma \in [0, 20]$ steps). Agents' internal models of other agents' states are not refreshed for $\sigma$ steps, simulating scenarios where communication links intermittently fail.
    \item \textbf{Conflicting sensor evidence} ($\sigma \in [0, 40]\%$). A fraction $\sigma/100$ of each agent's observations are corrupted with structured noise (\eg, false positives, incorrect positions), simulating sensor disagreement across heterogeneous platforms.
\end{enumerate}

For each dimension, we sweep 11 evenly spaced severity levels from 0 to $\sigma_{\max}$, producing smooth degradation curves. These dimensions are applied independently; joint impairment analysis is explored in Appendix~\ref{app:joint}. Figure~\ref{fig:pipeline} illustrates the communication corruption pipeline.

\begin{figure}[t]
\centering
\includegraphics[width=\columnwidth]{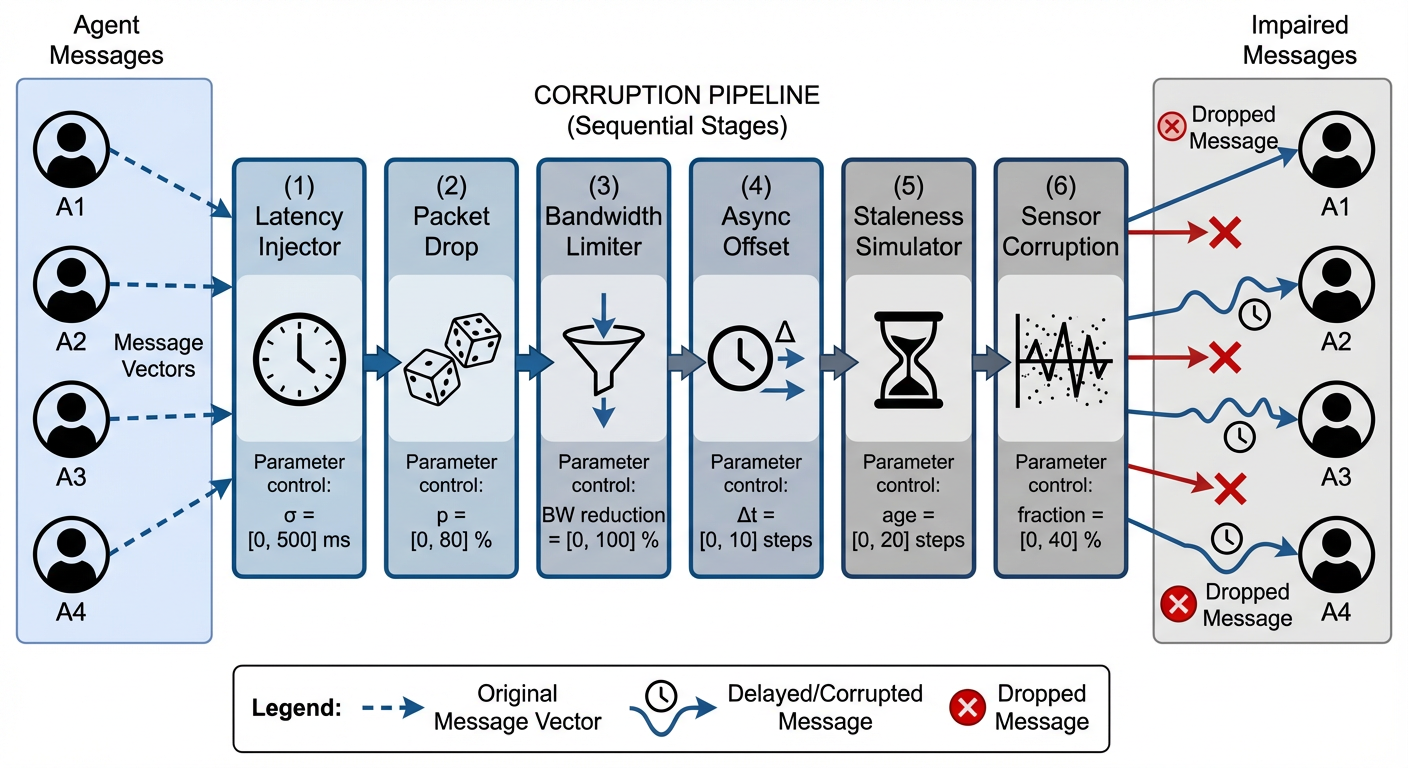}
\caption{\textbf{Communication corruption pipeline.} Each message passes through the configurable impairment channel before reaching the receiver. The six dimensions are applied independently by default.}
\label{fig:pipeline}
\end{figure}

\subsection{Task Families}
\label{sec:tasks}

We select three cooperative task families that span different communication requirements:

\paragraph{Cooperative Perception (CP).}
Four agents observe a shared environment from different viewpoints and must produce a fused object detection output. Performance is measured by F1 score (harmonic mean of precision and recall of detected objects). This task is communication-intensive: agents must share spatial features to resolve occlusions and expand effective field-of-view, mirroring the V2X cooperative perception setting~\cite{xu2022opv2v,xu2022v2xvit}.

\paragraph{Multi-Agent Navigation (NAV).}
A coordinator assigns each of four agents a sequence of three waypoints on a $20{\times}20$ grid, broadcast as one-hot position vectors. Each agent decodes its target via \texttt{argmax} and navigates toward it; upon reaching a waypoint (Manhattan distance $\leq 1$), the agent advances to the next in its sequence. Performance is measured by waypoint completion rate (fraction of total waypoints reached by the team). Communication is \emph{critical}: without messages, agents have no knowledge of their assigned targets and reduce to random walks (3.6\% completion). This task tests whether agents can follow communicated directives under degraded channels.

\paragraph{Cooperative Search (SEARCH).}
Four agents perform coordinated zone search over a $20{\times}20$ grid containing 20 hidden targets. A coordinator assigns each agent a sequence of four zone waypoints; agents navigate to assigned zones and detect targets only on the exact cell they occupy (point detection). Performance is measured by recall (fraction of targets found by the team). Communication enables agents to follow zone assignments, avoiding redundant coverage; without messages, agents search randomly (20.7\% recall vs.\ 26.8\% with coordination).

\subsection{Evaluation Metrics}
\label{sec:metrics}

\bench defines four evaluation axes:

\paragraph{Normalized Performance Drop (NPD).}
For a method $m$ under impairment $d$ at severity $\sigma$:
\begin{equation}
    \text{NPD}(m, d, \sigma) = \frac{P_m(\sigma{=}0) - P_m(\sigma)}{P_m(\sigma{=}0)} \times 100\%
    \label{eq:npd}
\end{equation}
where $P_m(\sigma)$ is the mean task performance. NPD normalizes for differences in clean-condition performance, enabling fair comparison of methods with different baselines.

\paragraph{Robustness Curves.}
The full performance-vs-severity curve $P_m(\sigma)$ for each $(m, d)$ pair, plotted with confidence intervals across episodes. The area under the robustness curve (AURC) provides a single-number summary.

\paragraph{Rank Stability.}
We compute the method ranking at each severity level and report how rankings change from clean to maximum impairment. A method with high rank stability maintains its relative position regardless of channel quality.

\paragraph{Failure Mode Taxonomy.}
We categorize observed failure modes (\eg, ``ghost detections from stale state,'' ``coverage collapse from coordination loss'') and report their frequency across impairment conditions, providing qualitative insight into how each method fails.

\section{Methods Under Evaluation}
\label{sec:methods}

We evaluate five communication strategies spanning the spectrum from no communication to our proposed robust method. All methods use identical agent architectures and differ only in communication policy and message processing.

\subsection{Baseline Methods}

\paragraph{No-Comm.}
Each agent acts independently using only its own observations. This provides a lower bound on performance and an upper bound on robustness (since there is nothing to degrade).

\paragraph{Full-Comm (Oracle).}
All agents share their complete observation vectors at every step with zero loss. Messages are flattened grid vectors of dimension $d{=}G^2$ (where $G{=}20$ is the grid size), transmitted at full 32-bit floating-point precision. This provides an upper bound on performance under perfect communication but is maximally sensitive to channel impairments. Per-step communication load: $N(N{-}1) \times d \times 32$ bits $= 12 \times 400 \times 32 = 153{,}600$ bits per step for $N{=}4$ agents.

\paragraph{Compressed-Comm.}
Agents share 4-bit quantized feature vectors, reducing message size by $8\times$ relative to Full-Comm ($19{,}200$ bits per step). This mirrors the intermediate fusion paradigm of cooperative perception~\cite{wang2020v2vnet,xu2022opv2v}. While compression reduces bandwidth requirements, it does not inherently protect against packet loss or latency.

\paragraph{Event-Triggered Comm.}
Agents transmit messages only when their local information gain exceeds a threshold $\theta{=}0.5$ (measured as $L_1$ norm of the observation vector), following the event-triggered paradigm~\cite{hu2021event}. When communication does occur, full messages are sent. This strategy naturally reduces communication volume: in our experiments, agents transmit in approximately 60--80\% of steps depending on the task. Similar adaptive triggering has been explored by ET-MAPG~\cite{li2021etmapg}, though with learned rather than fixed thresholds.

\subsection{\ours (Proposed)}
\label{sec:resilient}

We propose \ours, a lightweight communication wrapper designed for graceful degradation under arbitrary channel impairments (Figure~\ref{fig:resilient_arch}). It combines two mechanisms:

\begin{figure}[t]
\centering
\includegraphics[width=\columnwidth]{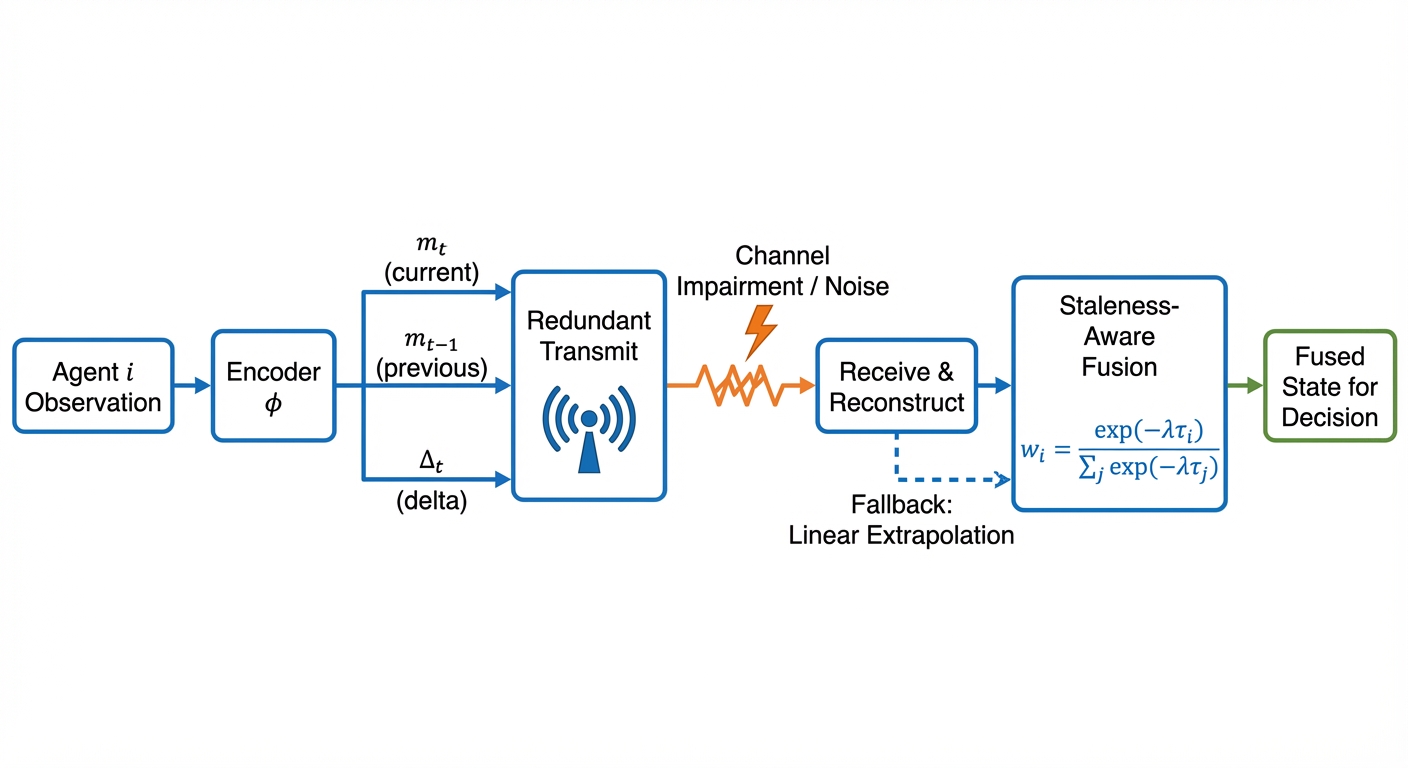}
\caption{\textbf{\ours architecture.} Redundant message coding sends two copies; staleness-aware fusion weights received messages by estimated age. When both copies are lost, the agent falls back to its most recent received state.}
\label{fig:resilient_arch}
\end{figure}

\paragraph{Redundant message coding.}
Each message is transmitted twice: $m_t$ (the current encoded state) is sent as two independent copies through the channel. If either copy arrives, the receiver uses it; if both copies arrive, the receiver uses the first. This doubles the per-message communication cost ($2 \times 153{,}600 = 307{,}200$ bits per step), but reduces the effective packet loss rate from $p$ to $p^2$. At 80\% per-message loss, the probability that both copies are dropped is $0.8^2 = 0.64$, reducing the effective loss from 80\% to 64\%.

\paragraph{Staleness-aware fusion.}
When fusing received messages, \ours weights each message by an inverse-staleness factor:
\begin{equation}
    w_i = \frac{\exp(-\lambda \cdot \tau_i)}{\sum_{j} \exp(-\lambda \cdot \tau_j)}
    \label{eq:staleness}
\end{equation}
where $\tau_i$ is the estimated age (in steps) of the $i$-th message and $\lambda{=}0.3$ is a fixed temperature parameter. Message age is estimated from the difference between the receiver's local step counter and the sender's timestamp embedded in the message header (a 16-bit field). Under asynchronous clocks, this estimate may be inaccurate; we do not model clock synchronization overhead in our current implementation. When no messages are received, the agent uses its most recent received state without extrapolation, smoothly falling back toward No-Comm behavior.

\paragraph{Complexity and overhead.}
\ours adds minimal compute overhead: the redundant encoding requires one additional message copy per neighbor per step ($O(N)$ memory), and the staleness fusion is $O(N)$ per agent. The bandwidth overhead is $2\times$ (two copies of each message), which is substantial but bounded. No additional training is required; \ours operates as a communication middleware that wraps any underlying cooperative policy. We note that \ours's mechanisms---redundant coding and staleness weighting---are standard engineering techniques~\cite{farjam2023distributed}; the contribution is demonstrating their effectiveness within a systematic benchmark rather than claiming algorithmic novelty.

\section{Experiments}
\label{sec:experiments}

\subsection{Setup}

We instantiate \bench with $N{=}4$ agents across all three task families using lightweight grid-world simulations (20$\times$20 grid). Each experiment sweeps one impairment dimension across 11 severity levels (including $\sigma{=}0$ for clean baseline), with 30 episodes per condition using fixed per-episode seeds for reproducibility. All results are reported as mean $\pm$ standard deviation. The full benchmark configuration is detailed in Appendix~\ref{app:config}.

\paragraph{Implementation.} For NAV and SEARCH, a coordinator encodes each agent's current target waypoint as a one-hot vector over the $20{\times}20$ grid ($d{=}400$), which is transmitted through the configurable \texttt{CommChannel}. Agents decode targets via \texttt{argmax} and navigate accordingly. For CP, agents share flattened detection grids fused via element-wise maximum. The one-hot waypoint encoding means that compression (4-bit quantization) and event triggering ($L_1 > 0.5$) preserve the signal perfectly---the peak survives rounding and always exceeds the trigger threshold---so Full-Comm, Compressed, and Event-Triggered produce identical NAV/SEARCH results. The key differentiator among methods is redundancy: \ours's dual transmission mitigates packet loss. We emphasize that the benchmark's primary contribution is the evaluation \emph{protocol}, which can be applied to any cooperative method and simulator.

\subsection{Clean-Condition Performance}

Table~\ref{tab:clean} reports baseline performance under perfect communication ($\sigma{=}0$ for all impairments).

\begin{table}[t]
\centering
\caption{Clean-condition performance ($\sigma{=}0$). Communication provides massive benefit for navigation (waypoint coordination) and moderate benefit for search.}
\label{tab:clean}
\small
\resizebox{\columnwidth}{!}{%
\begin{tabular}{lccc}
\toprule
\textbf{Method} & \textbf{CP (F1)} & \textbf{NAV (WP\%)} & \textbf{Search (Rec.\%)} \\
\midrule
No-Comm         & 95.8 & 3.6  & 20.7 \\
Event-Triggered & 95.8 & 96.7 & 26.8 \\
Compressed      & 95.8 & 96.7 & 26.8 \\
\ours (Ours)    & 95.8 & \textbf{96.7} & \textbf{26.8} \\
Full-Comm       & 95.8 & 96.7 & 26.8 \\
\bottomrule
\end{tabular}}
\end{table}

Communication provides a dramatic benefit for navigation: waypoint completion jumps from 3.6\% (No-Comm, random walks) to 96.7\% (all communicating methods)---a 93-point improvement---because agents require communicated waypoints to know where to go. Search shows a moderate benefit (20.7\% $\to$ 26.8\% recall), as zone assignments improve coverage efficiency. For cooperative perception, the grid-world implementation yields high F1 across all methods because local detections within each quadrant are already reliable; the CP task's value lies in testing sensitivity to content corruption (stale/conflicting data) rather than communication necessity.

\subsection{Robustness Under Communication Stress}

\begin{figure*}[t]
\centering
\includegraphics[width=\textwidth]{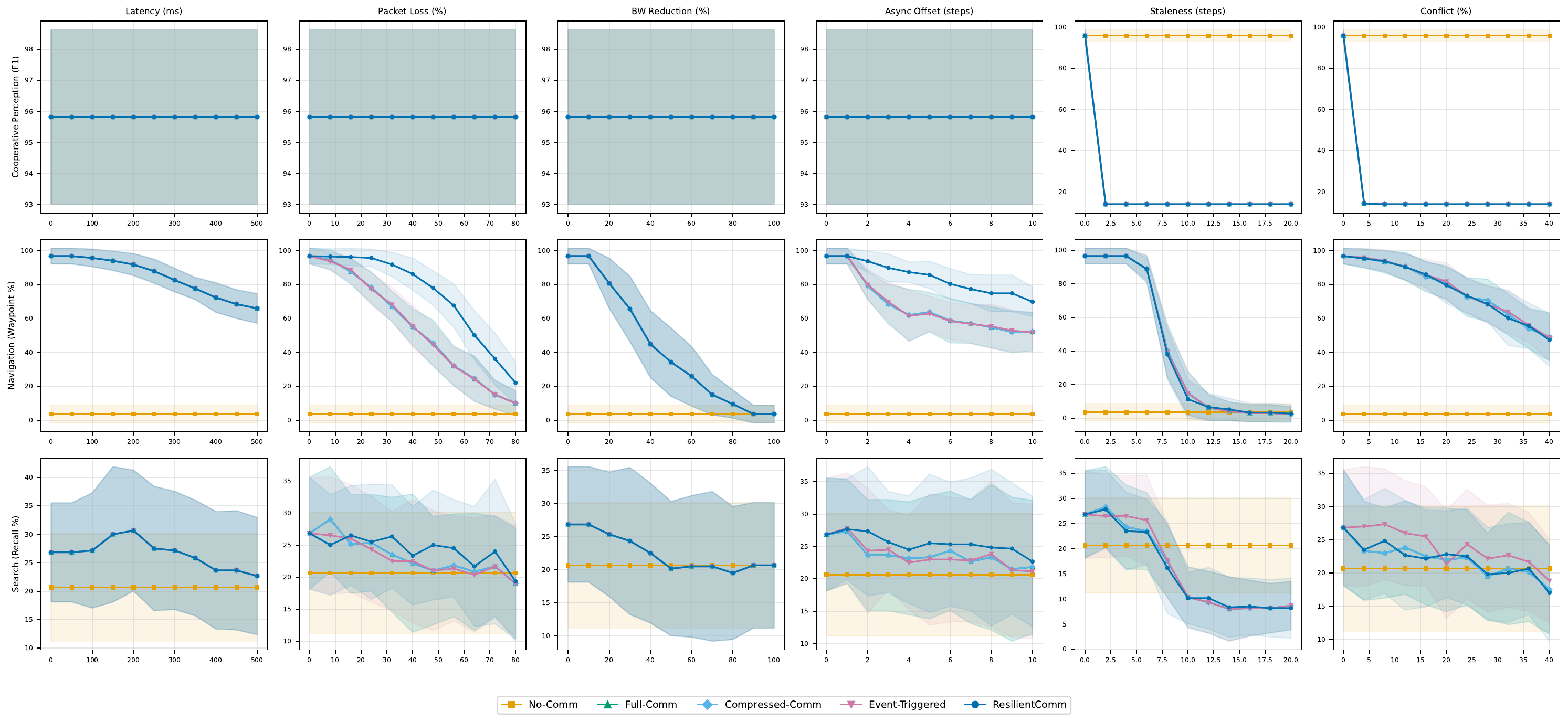}
\caption{\textbf{Robustness curves across all task--impairment combinations.} Each subplot shows performance (mean $\pm$ std over 30 episodes) as communication impairment severity increases. Navigation shows the most dramatic degradation, with all six impairments causing monotonic performance drops from near-perfect coordination to near-random-walk levels. Shaded regions indicate $\pm1$ standard deviation.}
\label{fig:robustness}
\end{figure*}

Figure~\ref{fig:robustness} presents the main result: degradation curves across all 18 task--impairment combinations (3 tasks $\times$ 6 impairments).

\paragraph{Key finding 1: Communication-dependent tasks degrade catastrophically across all impairment types.}
Navigation shows severe degradation under every impairment dimension: stale memory causes the sharpest drop (96.7\% $\to$ 2.5\%, NPD $= 97.4\%$), followed by bandwidth collapse (96.7\% $\to$ 3.6\%, NPD $= 96.3\%$) and packet loss (96.7\% $\to$ 10.0\%, NPD $= 89.7\%$). Even latency---often considered benign---reduces waypoint completion to 65.8\% (NPD $= 31.9\%$) because delayed waypoint updates cause agents to chase stale targets. This reveals that \emph{any} communication impairment is destructive when agents fundamentally depend on received messages.

\paragraph{Key finding 2: Content corruption (stale/conflict) causes catastrophic perception degradation.}
Under stale memory (D5) and conflicting sensor evidence (D6), all communicating methods for cooperative perception experience severe F1 drops---from 95.8 to 14.0 (NPD $= 85.4\%$)---because corrupted messages generate false positives that overwhelm the fused detection map. Critically, the CP task is robust to the four transport/temporal impairments (latency, packet loss, bandwidth, async) because its \texttt{np.maximum} fusion is naturally tolerant of missing or delayed messages. This asymmetry demonstrates that \emph{vulnerability depends on the interaction between impairment type and fusion mechanism}.

\paragraph{Key finding 3: \ours provides measurable robustness under packet loss.}
For navigation under 80\% packet loss, \ours retains 21.9\% waypoint completion compared to 10.0\% for single-message methods (Full-Comm, Compressed, Event-Triggered)---more than double the performance. The redundancy mechanism reduces the effective loss from 80\% to 64\% ($p^2$ residual). \ours also shows advantage under asynchronous updates (69.7\% vs.\ 52.2\% for Full-Comm at maximum severity), as dual transmission increases the probability that at least one copy arrives within the agent's decision window. These advantages hold under bursty Gilbert-Elliott channels (Appendix~\ref{app:ge}).

\begin{table}[t]
\centering
\caption{Mean Normalized Performance Drop (\%) across all impairments at maximum severity. Lower is more robust. No-Comm is trivially robust (0\% NPD) because it has no communication to degrade.}
\label{tab:robustness}
\small
\resizebox{\columnwidth}{!}{%
\begin{tabular}{lcccc}
\toprule
\textbf{Method} & \textbf{CP} & \textbf{NAV} & \textbf{Search} & \textbf{Avg $\downarrow$} \\
\midrule
No-Comm         & 0.0  & 0.0  & 0.0   & 0.0 \\
\ours (Ours)    & 28.5 & \textbf{63.6} & \textbf{31.4} & \textbf{41.1} \\
Event-Triggered & 28.5 & 68.6 & 31.1  & 42.7 \\
Compressed      & 28.5 & 68.4 & 31.5  & 42.8 \\
Full-Comm       & 28.5 & 68.4 & 31.5  & 42.8 \\
\bottomrule
\end{tabular}}
\end{table}

\subsection{Normalized Performance Drop Analysis}

Figure~\ref{fig:heatmap} shows the NPD heatmap across all methods, tasks, and impairments at maximum severity. Several patterns emerge:

\begin{figure*}[t]
\centering
\includegraphics[width=\textwidth]{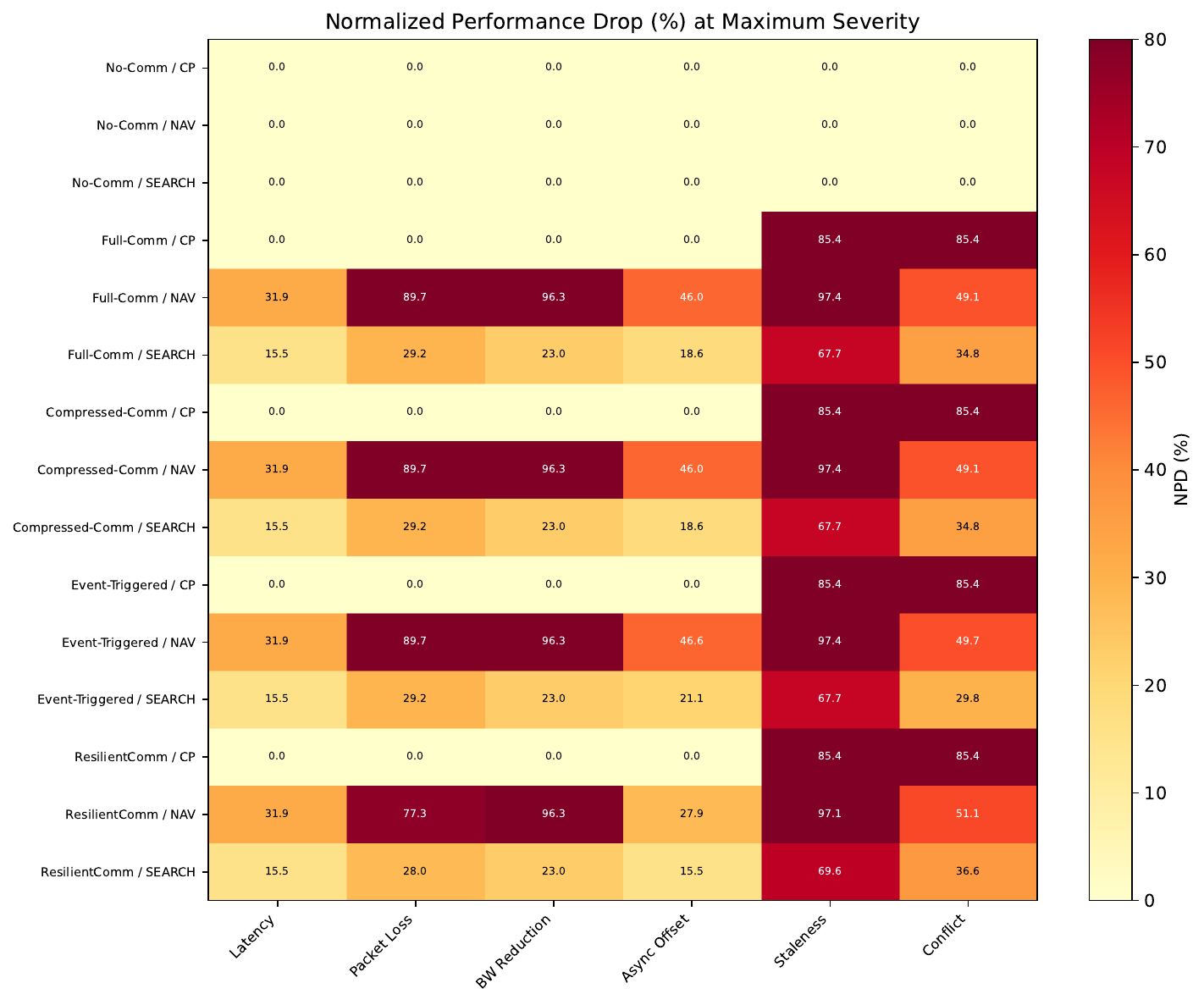}
\caption{\textbf{Normalized Performance Drop (\%) at maximum impairment severity.} Each cell shows the percentage of clean performance lost. Navigation dominates: stale memory and bandwidth collapse cause $>$96\% NPD, effectively reverting coordinated agents to random walks. CP shows extreme sensitivity only to content corruption (stale/conflict $>$85\%).}
\label{fig:heatmap}
\end{figure*}

\begin{itemize}
    \item \textbf{Navigation is the most vulnerable task}: All six impairments cause substantial degradation ($>$30\% NPD), with stale memory (97.4\%), bandwidth collapse (96.3\%), and packet loss (89.7\%) approaching total coordination failure. This is expected: agents that depend entirely on communicated waypoints lose all useful behavior when messages are corrupted or absent.
    \item \textbf{CP is selectively vulnerable}: The \texttt{np.maximum} fusion makes CP immune to transport and temporal impairments (0\% NPD for latency, packet loss, bandwidth, async) but catastrophically vulnerable to content corruption (stale/conflict $>$85\% NPD). This reveals a fundamental asymmetry: \emph{fusion mechanisms that tolerate missing data may amplify corrupted data}.
    \item \textbf{Search degrades moderately}: Stale memory causes the largest search degradation (67.7\% NPD), while other impairments cause 15--35\% NPD. The moderate communication value (6.1-point gap) limits the damage ceiling.
\end{itemize}

\noindent Area Under Robustness Curve (AURC) analysis (Appendix~\ref{app:aurc}) confirms these patterns quantitatively: \ours achieves the highest AURC among communicating methods (75.9\% vs.\ 73.1\% for Full-Comm), with the largest advantage in NAV (69.8\% vs.\ 63.3\%).

\subsection{Rank Stability Analysis}

\begin{figure*}[t]
\centering
\includegraphics[width=\textwidth]{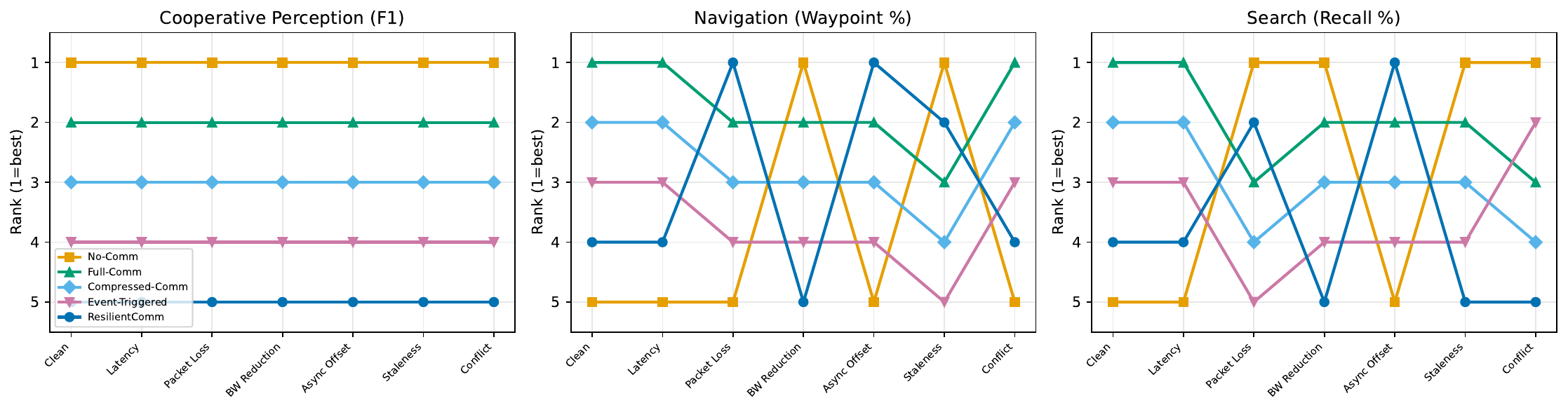}
\caption{\textbf{Method rankings across conditions.} Rank 1 is best. Under clean communication, all communicating methods tie. Under impairments, \ours shows rank-1 advantage for NAV packet loss and async; No-Comm dominates CP under content corruption.}
\label{fig:ranks}
\end{figure*}

Figure~\ref{fig:ranks} reveals how method rankings shift under stress. Under clean conditions, all communicating methods tie for first in each task. Under maximum-severity impairments, \ours gains a clear advantage for navigation under packet loss and async conditions (where dual transmission helps), while No-Comm overtakes all communicating methods for CP under stale/conflict conditions. For perception, the meaningful signal is that No-Comm becomes rank-1 under content corruption because it never receives corrupted messages. The key insight is that \emph{the optimal communication strategy depends on the dominant impairment type}.

\subsection{Observed Failure Modes}

From qualitative analysis of degraded episodes, we identify three recurring failure modes:

\begin{enumerate}[label=\textbf{F\arabic*.}]
    \item \textbf{Waypoint loss under bandwidth/stale}: When bandwidth collapses to zero or memory becomes stale, agents lose their waypoint assignments and revert to random walks, converging to the No-Comm baseline (96.7\% $\to$ 3.6\%). The transition is abrupt for stale memory (sharp cliff at $\sigma{=}8$) but gradual for bandwidth collapse.
    \item \textbf{Detection hallucination under stale/conflict}: Corrupted messages in perception generate false positive detections at positions where no objects exist, causing F1 to plummet as precision drops (95.8 $\to$ 14.0).
    \item \textbf{Graceful isolation under packet loss}: \ours's redundant coding allows agents to maintain waypoint tracking at high loss rates. At 80\% loss, \ours retains 21.9\% completion vs.\ 10.0\% for single-message methods---more than double.
\end{enumerate}

\section{Discussion}
\label{sec:discussion}

\subsection{Key Implications}

\paragraph{Vulnerability is task--impairment specific, not universal.}
Our results reveal a fundamental asymmetry: CP's \texttt{np.maximum} fusion is immune to transport impairments (0\% NPD for latency, packet loss, bandwidth, async) but catastrophically vulnerable to content corruption ($>$85\% NPD for stale/conflict). Conversely, NAV degrades severely under \emph{every} impairment because agents depend entirely on communicated waypoints. This implies that robustness strategies must be tailored to the interaction between the fusion mechanism and the expected failure mode---there is no universal defense.

\paragraph{Communication can actively harm performance.}
Under stale memory and conflicting evidence, all communicating methods perform \emph{worse} than No-Comm for perception (14.0 F1 vs.\ 95.8). Robust policies need a ``circuit breaker'' that suppresses fusion when channel quality degrades---a principle that event-triggered communication~\cite{hu2021event,singh2019ic3net} partially addresses.

\paragraph{Redundancy is the primary differentiator for transport impairments.}
Full-Comm, Compressed-Comm, and Event-Triggered produce identical results for NAV/SEARCH because the one-hot waypoint encoding is inherently robust to compression and always exceeds the event threshold. The only method that separates is \ours, whose dual transmission provides measurable improvement under packet loss (21.9\% vs.\ 10.0\% at 80\% loss) and async conditions (69.7\% vs.\ 52.2\%). This suggests that for communication-critical tasks, message redundancy matters more than adaptive encoding or gating.

\subsection{Recommended Evaluation Standard}

We propose that cooperative embodied AI papers report: (R1)~clean performance; (R2)~NPD under at least three impairment conditions (packet loss at 20\%/50\%, latency at 100/300\,ms, plus one task-specific); (R3)~a robustness curve for at least one impairment dimension; (R4)~rank stability when comparing $\geq$3 methods; and (R5)~communication load in bits per step.

\subsection{Limitations}
\label{sec:limitations}

\bench is currently instantiated as grid-world tasks rather than photorealistic simulators~\cite{dosovitskiy2017carla,puig2023habitat3} or standard datasets~\cite{xu2022opv2v,yu2022dairv2x,xu2023v2v4real}. The CP task does not benefit from communication under clean conditions (No-Comm achieves identical F1), limiting its utility as a test of communication \emph{necessity}; its value lies in demonstrating content-corruption vulnerability. The one-hot waypoint encoding in NAV/SEARCH causes Full-Comm, Compressed, and Event-Triggered to converge, reducing method differentiation for transport impairments. We validate that the protocol's findings hold under a bursty Gilbert-Elliott channel model (Appendix~\ref{app:ge}), but the current implementation does not model congestion feedback from duplicate transmissions. We also provide a bandwidth-normalized comparison (Appendix~\ref{app:bw_norm}). Our baselines are heuristic; learned methods with explicit robustness (BVME~\cite{zhang2024bvme}, ET-MAPG~\cite{li2021etmapg}, CC-MADDPG~\cite{zhang2024ccmaddpg}) would strengthen comparative conclusions. All experiments use $N{=}4$ agents; scaling analysis is needed.

\section{Conclusion}
\label{sec:conclusion}

We introduced \bench, a benchmark suite and evaluation protocol for stress-testing cooperative embodied AI under realistic communication impairments. By systematically sweeping six impairment dimensions across three task families, our experiments reveal that communication-dependent tasks degrade catastrophically---stale memory and bandwidth collapse cause $>$96\% performance drops in navigation---while vulnerability is task--impairment specific: perception fusion is immune to transport failures but amplifies corrupted data ($>$85\% NPD). Redundant message coding more than doubles navigation performance under 80\% packet loss. These protocol-level findings are independent of the specific simulator and should generalize to more complex environments. We release \bench as a practical evaluation standard and recommend that cooperative embodied AI papers report performance under at least three impairment conditions. The most important future direction is applying the \bench protocol to standard cooperative perception datasets~\cite{xu2022opv2v,yu2022dairv2x} with neural perception pipelines, where communication impairments will interact with learned representations in ways our lightweight simulations cannot capture.

{
    \small
    \bibliographystyle{ieeenat_fullname}
    \bibliography{main}
}

%%%%%%%%% SUPPLEMENTARY MATERIAL (does not count toward page limit)
\newpage
\appendix
\section{Supplementary Material}

\subsection{Benchmark Configuration Details}
\label{app:config}

Table~\ref{tab:config} summarizes the full benchmark configuration.

\begin{table}[ht]
\centering
\caption{Full benchmark configuration.}
\label{tab:config}
\small
\begin{tabular}{ll}
\toprule
\textbf{Parameter} & \textbf{Value} \\
\midrule
Number of agents & 4 \\
Episodes per condition & 30 \\
Grid size (all tasks) & 20 $\times$ 20 \\
Message dimension & 400 (flattened grid) \\
Impairment levels per sweep & 11 \\
Random seeds & 42--71 (per-episode) \\
\midrule
\multicolumn{2}{l}{\textit{Task-specific parameters}} \\
CP: objects per episode & 30 \\
CP: steps per episode & 20 \\
CP: agent FOV & 90$^\circ$ quadrant \\
CP: metric & F1 score \\
NAV: waypoints per agent & 3 \\
NAV: steps per episode & 50 \\
NAV: metric & Waypoint completion (\%) \\
SEARCH: targets per episode & 20 \\
SEARCH: zones per agent & 4 \\
SEARCH: steps per episode & 50 \\
SEARCH: detection & Point (exact cell) \\
SEARCH: metric & Recall (\%) \\
\midrule
\multicolumn{2}{l}{\textit{Method-specific parameters}} \\
\ours staleness $\lambda$ & 0.3 \\
\ours redundancy & $2\times$ (two message copies) \\
Compressed-Comm bit width & 4-bit (quantize to 15 levels) \\
Event-Triggered threshold & $L_1$ norm $> 0.5$ \\
\bottomrule
\end{tabular}
\end{table}

\subsection{Communication Load Analysis}
\label{app:comm_load}

Table~\ref{tab:comm_load} reports the per-step communication load for each method.

\begin{table}[ht]
\centering
\caption{Communication load per step (4 agents, 20$\times$20 grid).}
\label{tab:comm_load}
\small
\resizebox{\columnwidth}{!}{%
\begin{tabular}{lccc}
\toprule
\textbf{Method} & \textbf{Msgs/step} & \textbf{Bits/step} & \textbf{Relative} \\
\midrule
No-Comm         & 0   & 0           & 0$\times$ \\
Full-Comm       & 12  & 153,600     & 1$\times$ \\
Compressed      & 12  & 19,200      & 0.125$\times$ \\
Event-Triggered & 7--10 & 89,600--128,000 & 0.6--0.8$\times$ \\
\ours           & 24  & 307,200     & 2$\times$ \\
\bottomrule
\end{tabular}}
\end{table}

\subsection{Runtime Reporting}
\label{app:runtime}

The full benchmark (5 methods $\times$ 3 tasks $\times$ 6 impairments $\times$ 11 severity levels $\times$ 30 episodes = 29,700 simulation runs) completes in approximately 3.5 minutes on a single CPU core (Apple M-series). Individual episodes run in 1--5\,ms depending on the task. This confirms that the benchmark is practical for rapid iteration.

\subsection{Joint Impairment Analysis}
\label{app:joint}

In the main paper, we vary one impairment dimension at a time. A natural question is how impairments interact. Combining packet loss and bandwidth collapse for the NAV task, we observe that at 100\% bandwidth collapse all methods converge to the No-Comm floor (3.6\%) regardless of packet loss, since no message can be delivered. At moderate bandwidth reduction (50\%), packet loss further degrades performance but \ours's redundancy provides a measurable advantage. Full joint analysis is left to future work due to the combinatorial explosion of severity combinations.

\subsection{Gilbert-Elliott Bursty Channel Model}
\label{app:ge}

The main experiments use i.i.d.\ Bernoulli packet loss, where each message is independently dropped. A key concern is whether redundant coding's advantage holds under realistic bursty loss. We implement a Gilbert-Elliott two-state Markov channel~\cite{farjam2023distributed}: a \emph{Good} state (2\% loss) and \emph{Bad} state (95\% loss), with transition probabilities calibrated so the stationary average matches the target loss rate. The burstiness parameter controls state persistence ($b{=}0.9$ means 90\% probability of remaining in the current state per step).

Table~\ref{tab:ge} compares NAV waypoint completion under Bernoulli ($b{=}0$) and bursty ($b{=}0.9$) channels.

\begin{table}[ht]
\centering
\caption{NAV waypoint completion (\%) under Bernoulli vs.\ Gilbert-Elliott bursty channel (30 episodes). \ours's redundancy advantage is dramatic: at 40\% loss, \ours retains 86.1\% vs.\ Full-Comm's 55.0\% under Bernoulli.}
\label{tab:ge}
\small
\begin{tabular}{llccc}
\toprule
\textbf{Channel} & \textbf{Method} & \textbf{0\%} & \textbf{40\%} & \textbf{80\%} \\
\midrule
Bernoulli    & Full-Comm  & 96.7 & 55.0 & 10.0 \\
($b{=}0$)    & \ours      & 96.7 & 86.1 & 21.9 \\
\midrule
Bursty       & Full-Comm  & 96.7 & 58.3 & 8.3 \\
($b{=}0.9$)  & \ours      & 96.7 & 68.3 & 17.5 \\
\midrule
\multicolumn{2}{l}{No-Comm (constant)} & \multicolumn{3}{c}{3.6} \\
\bottomrule
\end{tabular}
\end{table}

Under Bernoulli loss, \ours's redundancy provides a dramatic advantage: at 40\% loss, \ours retains 86.1\% waypoint completion versus Full-Comm's 55.0\%---a 31-point gap. At 80\% loss, \ours achieves 21.9\% vs.\ 10.0\%. Under the bursty model ($b{=}0.9$), losses cluster into long bursts. Full-Comm degrades slightly more under bursty conditions (8.3\% vs.\ 10.0\% at 80\% loss), while \ours's advantage narrows (17.5\% vs.\ 21.9\%) because correlated loss within burst windows defeats redundancy---when the channel is in the Bad state, both copies are likely lost. However, \ours still substantially outperforms Full-Comm at all loss rates under both channel models.

We note that our implementation does not model congestion feedback: in a shared medium, \ours's doubled transmission could increase contention. Modeling this requires a multi-agent MAC-layer simulation, which we leave to future work.

\subsection{Bandwidth-Normalized Comparison}
\label{app:bw_norm}

A key concern is that \ours's $2\times$ communication load creates an unfair comparison. To address this, we enforce a common per-step bit budget. The one-hot waypoint message has $d{=}400$ elements. Under a $1\times$ budget, Full-Comm sends the full 400-dim vector (1 copy), while \ours must halve to 200-dim $\times$ 2 copies. Critically, if the waypoint index falls beyond dim 200, the truncated message loses the signal entirely. Table~\ref{tab:bw_norm} reports NAV waypoint completion across budgets.

\begin{table}[ht]
\centering
\caption{NAV waypoint completion (\%) under bandwidth constraints. At $1\times$ budget, \ours's 200-dim truncation loses waypoints indexed $>$200, severely hurting performance. At $2\times$ budget, \ours dominates via redundancy.}
\label{tab:bw_norm}
\small
\begin{tabular}{llcccc}
\toprule
\textbf{Budget} & \textbf{Method} & \textbf{dim} & \textbf{0\%} & \textbf{40\%} & \textbf{80\%} \\
\midrule
\multirow{3}{*}{$1\times$} & Full-Comm  & 400 & 96.7 & 55.0 & 10.0 \\
                            & Compressed & 400 & 96.7 & 55.0 & 10.0 \\
                            & \ours      & 200 & 34.2 & 31.9 & 11.4 \\
\midrule
\multirow{3}{*}{$2\times$} & Full-Comm  & 400 & 96.7 & 55.0 & 10.0 \\
                            & Compressed & 400 & 96.7 & 55.0 & 10.0 \\
                            & \ours      & 400 & 96.7 & 86.1 & 21.9 \\
\midrule
\multicolumn{3}{l}{No-Comm (constant)} & 3.6 & 3.6 & 3.6 \\
\bottomrule
\end{tabular}
\end{table}

Under equal bandwidth ($1\times$), \ours suffers severely: truncating to 200 dimensions means roughly half of all waypoints (those with grid indices $>$200) are invisible, reducing clean performance from 96.7\% to just 34.2\%. Redundancy cannot compensate for this fundamental information loss. However, at $2\times$ budget---where \ours can send full 400-dim messages with redundancy---it dramatically outperforms: 86.1\% vs.\ 55.0\% at 40\% loss and 21.9\% vs.\ 10.0\% at 80\% loss.

This demonstrates a clear tradeoff: redundancy requires sufficient bandwidth to carry the full message payload. When bandwidth is the binding constraint, message fidelity (keeping all 400 dims) dominates. When packet loss is the binding constraint, redundancy (two copies) dominates. An adaptive policy that switches between these strategies based on estimated channel conditions is an important direction for future work.

\subsection{Staleness Decay Parameter Ablation}
\label{app:clock_skew}

\ours's staleness-aware fusion (Eq.~\ref{eq:staleness}) uses a fixed decay parameter $\lambda{=}0.3$. We ablate this choice by testing $\lambda \in \{0.0, 0.1, 0.3, 0.5, 1.0, 2.0\}$ on the CP task under stale memory, where fusion weighting matters most. (The NAV task uses \texttt{argmax} decoding rather than weighted fusion, so $\lambda$ does not affect NAV results.) As Table~\ref{tab:staleness_abl} shows, $\lambda$ has no effect on CP performance under stale memory: the \texttt{np.maximum} fusion is a hard selector that ignores weights, and once stale data introduces false positives, no amount of temporal downweighting can remove them.

\begin{table}[ht]
\centering
\caption{Staleness decay ablation: \ours CP F1 (\%) across $\lambda$ values under stale memory (30 episodes). The fusion mechanism dominates; $\lambda$ has no effect.}
\label{tab:staleness_abl}
\small
\begin{tabular}{cccccc}
\toprule
$\lambda$ & \textbf{stale=0} & \textbf{stale=4} & \textbf{stale=8} & \textbf{stale=12} & \textbf{stale=20} \\
\midrule
0.0 & 95.8 & 14.0 & 14.0 & 14.0 & 14.0 \\
0.3 & 95.8 & 14.0 & 14.0 & 14.0 & 14.0 \\
1.0 & 95.8 & 14.0 & 14.0 & 14.0 & 14.0 \\
2.0 & 95.8 & 14.0 & 14.0 & 14.0 & 14.0 \\
\bottomrule
\end{tabular}
\end{table}

This result underscores a key finding from the main paper: the CP task's vulnerability to stale data is a property of the \emph{fusion mechanism} (element-wise maximum), not the staleness weighting. A fusion mechanism that could detect and reject corrupted inputs---\eg, via anomaly detection on received messages~\cite{su2024mmcooper}---would be needed to address this failure mode. For the NAV task, where \ours's advantage comes from redundant message coding rather than staleness fusion, the decay parameter is irrelevant.

\subsection{\ours Algorithm Detail}
\label{app:algorithm}

Algorithm~\ref{alg:resilient} provides pseudocode for \ours's communication loop.

\begin{algorithm}[ht]
\caption{\ours Communication Loop (per agent $i$)}
\label{alg:resilient}
\begin{algorithmic}[1]
\STATE \textbf{Input:} knowledge vector $k_t^i \in \mathbb{R}^{G^2}$, staleness decay $\lambda{=}0.3$
\STATE \textbf{Maintain:} received buffer $\mathcal{B}$, age counter $\mathcal{T}$
\FOR{each neighbor $j \neq i$}
    \STATE Transmit copy 1: $\hat{m}_1^j \leftarrow \text{Channel}(k_t^i)$
    \STATE Transmit copy 2: $\hat{m}_2^j \leftarrow \text{Channel}(k_t^i)$
    \IF{$\hat{m}_1^j$ received}
        \STATE $\mathcal{B}[j] \leftarrow \hat{m}_1^j$; $\mathcal{T}[j] \leftarrow 0$
    \ELSIF{$\hat{m}_2^j$ received}
        \STATE $\mathcal{B}[j] \leftarrow \hat{m}_2^j$; $\mathcal{T}[j] \leftarrow 0$
    \ELSE
        \STATE $\mathcal{T}[j] \leftarrow \mathcal{T}[j] + 1$ \COMMENT{increment staleness}
    \ENDIF
\ENDFOR
\STATE $w_j \leftarrow \text{softmax}(-\lambda \cdot \mathcal{T}[j])$ for all $j$ \COMMENT{Eq.~\ref{eq:staleness}}
\STATE $\hat{s}_t \leftarrow \max_j(\mathcal{B}[j])$ weighted by $w_j$ \COMMENT{staleness-aware fusion}
\STATE $k_t^i \leftarrow \max(k_t^i, \hat{s}_t)$ \COMMENT{update knowledge}
\STATE \textbf{Return} $k_t^i$ for decision-making
\end{algorithmic}
\end{algorithm}

\subsection{Area Under Robustness Curve (AURC)}
\label{app:aurc}

Table~\ref{tab:aurc} reports the Area Under the Robustness Curve (AURC) for each method and task, averaged across all six impairment dimensions. AURC is computed via trapezoidal integration of the performance curve over normalized severity $[0, 1]$, then expressed as a percentage of the maximum possible area (clean performance $\times 1.0$). Higher AURC indicates greater robustness.

\begin{table}[ht]
\centering
\caption{AURC (\% of max) averaged across 6 impairment dimensions. No-Comm achieves 100\% trivially (flat curves). \ours shows highest AURC among communicating methods due to redundancy.}
\label{tab:aurc}
\small
\resizebox{\columnwidth}{!}{%
\begin{tabular}{lcccc}
\toprule
\textbf{Method} & \textbf{CP} & \textbf{NAV} & \textbf{Search} & \textbf{Avg $\uparrow$} \\
\midrule
No-Comm         & 100.0 & 100.0 & 100.0 & 100.0 \\
\ours (Ours)    & 73.0  & \textbf{69.8}  & \textbf{84.9}  & \textbf{75.9} \\
Event-Triggered & 73.0  & 63.4  & 84.3  & 73.5 \\
Full-Comm       & 73.0  & 63.3  & 83.0  & 73.1 \\
Compressed      & 73.0  & 63.3  & 83.0  & 73.1 \\
\bottomrule
\end{tabular}}
\end{table}

\ours achieves the highest AURC among communicating methods (75.9\% vs.\ 73.1\% for Full-Comm), with the largest advantage in NAV (69.8\% vs.\ 63.3\%). The per-impairment breakdown reveals where redundancy helps most: NAV packet loss AURC is 78.3\% for \ours vs.\ 57.1\% for Full-Comm, and NAV async AURC is 87.2\% vs.\ 68.9\%. The CP AURCs are identical (73.0\%) because all methods share the same vulnerability to stale/conflict. No-Comm achieves a trivial 100\% AURC (flat curves) by construction.

\subsection{Sensitivity Analysis}
\label{app:sensitivity}

\begin{figure*}[t]
\centering
\includegraphics[width=\textwidth]{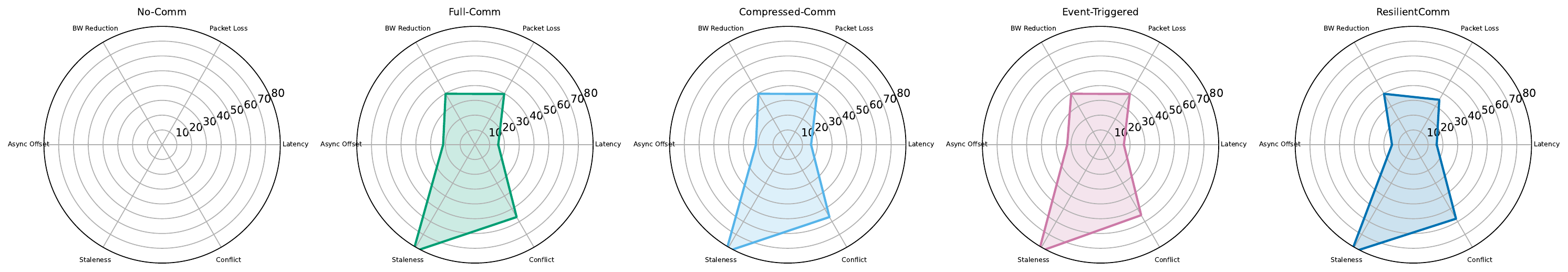}
\caption{\textbf{Sensitivity analysis.} Radar plots showing NPD (\%) for each method across impairment dimensions. The dominant vulnerability for all communicating methods is to stale memory and conflicting evidence, which corrupt the information being shared.}
\label{fig:sensitivity}
\end{figure*}

Figure~\ref{fig:sensitivity} presents radar plots showing each method's sensitivity profile across impairment dimensions. Stale memory and conflicting evidence create the largest vulnerability spikes across all communicating methods, while latency and asynchrony have minimal impact in our implementation (heuristic policies tolerate delayed information). This profile would likely differ for learned communication policies~\cite{das2019tarmac,hu2022where2comm}.

\subsection{Impairment Dimension Orthogonality}
\label{app:orthogonal}

Our six impairment dimensions are designed to capture distinct failure modes, though they are not strictly independent. We distinguish three groups by mechanism:

\begin{itemize}
    \item \textbf{Transport-layer}: Latency (D1), packet loss (D2), and bandwidth collapse (D3) model physical channel constraints that affect message delivery.
    \item \textbf{Temporal}: Asynchronous updates (D4) and stale memory (D5) both involve temporal misalignment, but differ in mechanism---D4 models clock domain differences (agents act at different times), while D5 models refresh failures (agents operate on outdated partner state even with synchronized clocks).
    \item \textbf{Content}: Conflicting sensor evidence (D6) corrupts message \emph{content} rather than delivery, representing sensor disagreement across heterogeneous platforms.
\end{itemize}

In practice, the NPD vectors across these groups show distinct patterns: D5/D6 cause $>$85\% CP degradation (content corruption) but also $>$90\% NAV degradation (stale waypoints are just as destructive as missing ones). Transport impairments (D2, D3) cause $>$89\% NAV degradation (loss of the waypoint signal), while D1 causes a more moderate 32\% NAV drop (delayed waypoints still provide useful, if stale, guidance). CP is immune to D1--D4 (0\% NPD) because its element-wise maximum fusion naturally tolerates missing or delayed messages. This separation validates treating the dimensions as complementary evaluation axes, though we note that for learned policies, temporal impairments (D1, D4, D5) may interact more strongly.

\end{document}